\pdfoutput=1

\documentclass[11pt]{article}

\usepackage{acl}

\usepackage{times}
\usepackage{latexsym}

\usepackage[T1]{fontenc}

\usepackage[utf8]{inputenc}

\usepackage{microtype}

\usepackage{multicol}
\usepackage{multirow}
\usepackage{graphicx}
\usepackage{enumitem}
\usepackage{color}

\title{SU-NLP at SemEval-2022 Task 11: Complex Named Entity Recognition with Entity Linking}

\author{Buse Çarık$^*$, Fatih Beyhan$^*$, Reyyan Yeniterzi \\
         Sabancı University \\ İstanbul, Turkey \\
         \texttt{\{busecarik,fatihbeyhan,reyyan\}@sabanciuniv.edu} }

\begin{document}
\maketitle
\def\thefootnote{*}\footnotetext{These authors contributed equally to this work.}
\begin{abstract}

This paper describes the system proposed by Sabancı University Natural Language Processing Group in the SemEval-2022 MultiCoNER task. We developed an unsupervised entity linking pipeline that detects potential entity mentions with the help of Wikipedia and also uses the corresponding Wikipedia context to help the classifier in finding the named entity type of that mention. Our results showed that our pipeline improved performance significantly, especially for complex entities in low-context settings.

\end{abstract} 

%
%
%
%
%

\section{Introduction}
\label{intro}

%


Named Entity Recognition (NER) is a widely studied task of Natural Language Processing. Recent leading architectures achieved impressive performances, especially in formal contexts (like news articles, etc.) and for commonly worked named entity types (like Person, Organization, etc.).
In \textit{Multilingual Complex Named Entity Recognition} task, organizers introduced a dataset which is a large collection of short texts that includes complex entities. As they stated, complex entities can be noun phrases, gerunds, or full clauses \cite{multiconer-data}. The participants were challenged to develop a NER system for identifying these complex entities in short and low-context settings.

Recent contextual neural architectures like transformers work quite well on a variety of NLP tasks and datasets. However, they are strongly dependent on the context of the given text. In case of lack of context, these models alone may not gather enough evidence to make a correct prediction. When that is the case, getting help from knowledge bases can be useful. 
Dataset provided with this task has similar low-context characteristics together with some complex named entity types to be resolved. Therefore, in this work, we investigated whether performing entity linking to provide additional context for a transformer model can be useful for identifying complex named entities. We initially used an unsupervised entity linking approach to identify the possible entity mentions and their corresponding Wikipedia pages. Later on, we used the additional context retrieved from Wikipedia to predict the types of the identified entities. Our results show that performing entity linking prior to the classifier is beneficial in the case of complex entities in short and low-context settings.


\section{Related Work}\label{related_work}
%
%
%

NER is an important and also well-studied task in NLP. 
Similar to other NLP tasks, recent neural architectures provided significant improvements in NER as well. However, since these models are heavily dependent on context information, applying them to short, noisy texts such as very short social media posts or web queries results in significant performance degradation \cite{meng2021gemnet}. 

To overcome the lack of context in these short texts, as well as the code-mixed terms, it has been prominent to use external knowledge in neural approaches. Recent studies \cite{meng2021gemnet, fetahu2021gazetteer} combining multilingual transformer models with gazetteers have shown remarkable improvement in performance.
Another study \cite{yamada2015enhancing} utilized the Entity Linking task to address the issues of Twitter NER by detecting entity mentions from knowledge bases like Wikipedia. 

In this study, we also focused on external information due to the lack of context. However, instead of using a static auxiliary like gazetteers, we tried to detect entities dynamically from the continuously extending Wikipedia by using a search engine architecture. We integrated the retrieved Wikipedia content into the input representation in order to improve the performance of entity type classification.  

\section{Data} 

\citet{multiconer-data} introduced a new multilingual NER dataset consisting of short texts for eleven languages. The dataset consists of 6 named entity classes which are Person, Location, Group, Corporation, Product, and Creative Work. 

In this paper, even though we propose a language-independent approach, due to limited resources we focused on the Turkish part of the dataset only. Statistics of the dataset for Turkish are shown in Table \ref{tab:data_stats}. As can be seen from Table \ref{tab:data_stats}, even though the number of samples is much higher, the length of the samples is significantly lower in the test set compared to train and validation splits. This short length in context may cause recent state-of-the-art contextual models to perform worse than expected.


\begin{table}[!ht]
\begin{center}
\begin{tabular}{l|r|r}
     & \# Samples & Avg. \# Tokens \\ \hline
Train & 15,300 & 14.27  \\
Validation & 800 & 14.27  \\
Test & 136,935 & 5.28 \\
\end{tabular}
\caption{Distribution of samples and average number of tokens per sample for Turkish part}
\label{tab:data_stats}
\end{center}
\end{table}





An issue we found with the dataset is that in some examples, a phrase is labeled as \textit{OTHER} although there are similar phrases that are labeled as a named entity. Consider the following example:

\noindent \textit{"{\color{red}Nokia 1011}, 1994'e kadar, {\color{blue}Nokia 2010} ve {\color{blue}Nokia 2110}'un halefler olarak tanıtılmasıyla üretime devam etti."
    \textit{English: "{\color{red}Nokia 1011} continued in production until 1994, with the introduction of the {\color{blue}Nokia 2010} and {\color{blue}Nokia 2110} as successors.}"}

The text includes three different cellphone models produced by Nokia. Two of these models (the blue ones) were annotated as \textit{PRODUCT}; however, \textit{Nokia 1011} was mislabeled as \textit{OTHER}. 
A more major issue is when the same named entity occurs more than once in a sentence, but their annotations are different. An example is illustrated below:

\noindent \textit{"{\color{blue}Whitney Houston} yine bu sene içinde New Jack Swing'e adını yazdırdı, I'm Your Baby Tonight adlı şarkı {\color{red}Whitney Houston}'a başarı üstüne başarı getirmiştir." English: "\textit{{\color{blue}Whitney Houston} made her name on New Jack Swing again this year, the song I'm Your Baby Tonight brought {\color{red}Whitney Houston} success after success."}}

In the example, Whitney Houston in blue was labeled as \textit{PERSON}; but, the red Whitney Houston was labeled as \textit{OTHER}. These types of wrong annotations cause ambiguities and learning problems in supervised approaches.  

\section{Methodology}\label{methodology}
\label{sec:methodology}
%

In order to overcome the challenges described in previous sections, we propose an unsupervised entity linking pipeline followed by a classifier as our proposed NER architecture. The entity linking pipeline is designed to detect potential entity mentions and retrieve corresponding candidate documents for each mention. Later, these detected mentions and the linked content is used together in a classifier to detect the named entity type of the entity mention.
A pre-trained BERT model was also fine-tuned for the Named Entity Recognition task in order to create a strong baseline for comparison.


\begin{figure*}
    \includegraphics[width=\textwidth]{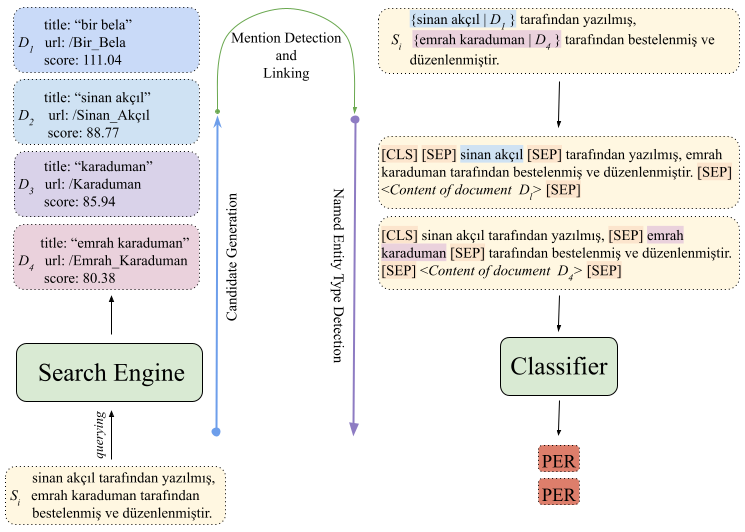}
    \caption{Overall pipeline of the system which is showing the path of sentence \textit{$S_i$}.}
    \label{fig:pipeline}
\end{figure*}

\subsection{BERT}\label{ner_bert}

The BERT \cite{BERT} model was adopted as a baseline approach since transformer models outperform other approaches in many NLP tasks. We fine-tuned the BERTurk \cite{BERTurk} model pre-trained on large Turkish corpora with a feed-forward layer on top to classify named entities. Base model, in which each feed-forward layer has 12 encoder layers and 768 hidden units, was used. Since every query in all datasets consists of lowercase letter, uncased version of BERTurk was preferred. 

\subsection{NER with Entity Linking}\label{es_bert}

BERT like transformer models heavily depend on the context. Therefore having a very short context like a couple of words, not even a sentence will cause problems with these models. Especially when the task is to identify complex named entities like creative work or product, it is much more challenging. For very short context, even human beings may have hard time resolving entities and their types. 

In order to overcome this challenge an additional help from a gazetteer or a knowledge base will be very useful. Therefore in this paper we propose using Wikipedia in order to both identify entities and also their types. Our proposed architecture which is depicted in Figure \ref{fig:pipeline} consists of three steps. The first two steps work towards identifying possible entities and linking them to their corresponding Wikipedia articles. In this unsupervised entity linking approach as the first step the original input text is used to retrieve relevant Wikipedia pages which are possible candidate pages for entities. In the second step content within the retrieved documents (like their titles) are used to identify the entity mentions in the original text. Finally, in the third step each identified mention and its linked Wikipedia article is used together to predict the type of the entity.

\subsubsection{Candidate Generation}\label{candidate_generation}
ElasticSearch\footnote{https://www.elastic.co/} was used as the underlying search engine architecture. The most recent Wikipedia dump was used to create an index. For each document \textit{$D_i$}, the \textit{title}, \textit{referred\_by} and \textit{interwikies} were indexed to create the following four fields:

\begin{itemize}
\item{\textit{title}:} This is the title of the document \textit{$D_i$}.
\item{\textit{referred\_by}:} This is a set of text spans which are parsed from other Wikipedia pages where \textit{$D_i$} is being referred with a interwiki link. This set includes all of these text spans sorted from longest to shortest. The title is added to the set as well.
\item{\textit{interwikies}:} This is the list of interwikies\footnote{Interwikies are the types of hyperlinks which is linking a text span from a Wikipedia page to another Wikipedia page. For instance, Wikipedia pages of Earth, Mercury and Venus are in the interwikies section of the Sun's Wikipedia page, since they are mentioned within its text.} in the document \textit{$D_i$}.
\item{\textit{all\_text}:} This one is a concatenation of all fields (title, text content, interwikies, referred\_by, and categories\footnote{Categories section of a Wikipedia page includes a list of topics which is related to the Wikipedia page.}) of the document \textit{$D_i$}. 
\end{itemize}



Each sample was queried and the most relevant 200 articles were retrieved. Later on for each query, documents retrieved with different field searches were pooled together. The reason we are pooling these different field results is to make sure we do not miss a possibly relevant article which could have been linked.


For the train and validation tests, around 199 documents were retrieved for each field. After pooling, we ended up with a pool of size 578 documents on average. Since test queries are shorter, the average number of documents returned got as low as 161 with the title field and 465 over the pool. Short queries also affected the empty ones, in other words, no document retrieved samples. After pooling there was not such a case for train and validation, however even pooling could not help to the 169 queries over 136,935 samples, and no Wikipedia articles were retrieved for those.




\begin{table}[]
    \centering
    \begin{tabular}{l|c c}
         Field & Train & Validation \\ \hline
         \textit{title}       & 96.52 & 97.00 \\ 
         \textit{interwikies} & 57.16 & 61.26 \\
         \textit{referred\_by}& 96.46 & 96.11 \\
         \textit{all\_text}   & 86.50 & 86.68 \\\hline
         After pooling        & 98.89 & 98.85 
    \end{tabular}
    \caption{Recall Scores over Detected Mentions}
    \label{tab:entity_detection_table}
\end{table}

\subsubsection{Mention Detection and Linking}\label{mention_detection_linking}

After generating a set of candidate documents for each input text, the next step is to map these documents to the possible entity mentions in the input. For each input, our proposed algorithm iterated over the retrieved and pooled documents \{\textit{D}\}, and cross-checked each item in the \textit{referred\_by} field of the document to see if there was an exact match to any phrase in the input text. In case there was an exact match, then that matched phrase was tagged as a possible entity and linked to the corresponding Wikipedia page. If there were multiple matches for a span of text, then the longest match was considered and shorter ones were ignored. Furthermore, if there was a tie in terms of length, then the relevance score of the Wikipedia document was used. That relevance score was calculated after pooling by doing a summation of the relevance scores from different field retrievals.  

In order to see which fields in Wikipedia are more useful for mention detection, recall was calculated over detected entity mentions, without considering the type of entity. The results are summarized in Table 2. Based on the results \textit{title} and \textit{referred\_by} are the most useful fields and pooling all retrievals is the best over both train and development sets.  




\subsubsection{Named Entity Type Detection}\label{named_entity_type_detection}

After linking the entities, the only thing remaining is to find their NER type. In order to do that we used the original context where the entity was mentioned in the input text and the linked Wikipedia article content. We combined these two as follows:
\begin{itemize}[leftmargin=0in]
    \item[] {\fontfamily{qcr}\selectfont[CLS]}\textit{ctx\textsubscript{l}}{\fontfamily{qcr}\selectfont[SEP]}\textit{M}{\fontfamily{qcr}\selectfont[SEP]}\textit{ctx\textsubscript{r}}{\fontfamily{qcr}\selectfont[SEP]}\textit{WP}{\fontfamily{qcr}\selectfont[SEP]}
\end{itemize}
where \textit{ctx\textsubscript{l}} and \textit{ctx\textsubscript{r}} are tokens for left and right context of mention, \textit{M} is mention itself and \textit{WP} is  the first two paragraphs of a Wikipedia page. {\fontfamily{qcr}\selectfont[SEP]} tokens are used to tag the mention. The maximum length of the representation is set 256. This representation is given to BERT for encoding. 

\begin{figure}
\begin{center}
    \includegraphics[scale=0.45]{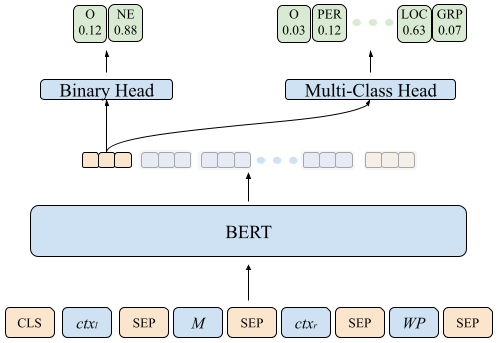}
    \caption{The architecture of the classifier. NE in the binary classifier head's output stands for \textit{Named Entity} and O for \textit{Other}.}
    \label{fig:multi_head}
\end{center}
\end{figure}

We used two classifier heads following the BERT encoder for classifying the pairs. The architectural design is presented in Figure \ref{fig:multi_head}. The first one is a binary classifier head which aims to solve an auxiliary task of whether the given candidate entity mention is a real entity or not with the help of input context tokens and Wikipedia content. The second classifier head is trying to learn the type of the candidate named entity. The loss function for our model is the summation of the two losses from these classifier heads. In case the binary classifier head returns O (which means the given candidate entity mention is not an entity), the second classifier head's decision is ignored, and given candidate entity mention is directly predicted as O (which stands for Other, not an entity).
In order to test the effectiveness of multiple heads, we created another model which has a single classifier head only for identifying named entity types.  

\section{Experimental Setup}\label{experimental_settings}

%

\begin{table*}[!ht]
\begin{center}
\begin{tabular}{l|ccc|ccc}
    \multirow{2}{*}{Models} & \multicolumn{3}{c|}{Validation} & \multicolumn{3}{c}{Test} \\
                           & Precision & Recall & F1 & Precision & Recall & F1 \\\hline
    BERT & 85.84 & 85.01 & \textbf{85.39} & 58.02 & 58.39 & 57.82 \\
    \hline
    EL\_BERT & 84.59 & 79.40 & 81.74 & 74.78 & 59.03 & 65.47 \\
    EL\_MultiBERT & \textbf{86.18} & 77.56 & 81.60 & \textbf{80.69} & 65.13 & 71.91 \\ \hline
    Ensemble$_{Token+EL}$ & 83.78 & 85.35 & 84.47 & 74.23 & 59.44 & 65.66 \\ 
    Ensemble$_{Token+EL\_Multi}$ & 83.66 & \textbf{85.51} & 84.49 & 78.86 & \textbf{66.43} & \textbf{72.02} \\
\end{tabular}
\caption{Results of our proposed methods on development and test data}
\label{tab:model_results}
\end{center}
\end{table*}

Due to limited resources and time, we only experimented with the Turkish part of the dataset and therefore indexed 749,204 Turkish Wikipedia pages.
During the retrieval \textit{OR} operator was used as part of the query and the default BM25 was used for ranking.

We fine-tuned the BERTurk model\footnote{https://huggingface.co/dbmdz/bert-base-turkish-uncased}. As the optimizer, we used AdamW \cite{loshchilov2017decoupled} with a fixed learning rate in all our BERT models as 3 x $10^{-5}$. We also applied dropout before the feed-forward layer with 0.3 probability. 
Our batch size was 32, with a maximum sequence length of 128 for the baseline BERT. We extended the maximum sequence length to 256 for EL\_MultiBERT and reduced the batch size to 8. All experiments were conducted with seed 22.

%
%
%
%

%

\section{Results}\label{sec:results}

The official evaluation metric was chosen as macro-average F1-score by the organizers. In Table \ref{tab:model_results}, we present our results with models described in Section \ref{sec:methodology}. 
Our baseline BERT, the one fine-tuned directly for NER, worked quite well on the validation set but performed badly on the test set. This substantial difference in these two sets shows that contextual models like BERT do not perform well in low-context settings.

On the other hand, although our proposed entity linking approaches fell slightly behind the baseline model in the development set, they performed significantly better than the baseline BERT in testing.
Among these two entity linking methods, the model with the two classification heads performed better. While determining the named entity type, training with this auxiliary task even using the same data seems to be useful. 

\begin{table}[]
    \centering
    \begin{tabular}{l|c|c}
        Entity Type & BERT & EL\_MultiBERT \\ \hline
        Person & 70.72 & 72.43 \\
        Location & 59.73 & 57.42 \\
        Group & 51.49 & 76.51 \\
        Corporation & 57.69 & 73.79 \\
        Product & 63.13 & 76.93 \\
        Creative Work & 44.14 & 74.40 \\
    \end{tabular}
    \caption{F1-Scores by NER type on Test Set}
    \label{tab:ne_dist_results}
\end{table}

Table \ref{tab:ne_dist_results} illustrates the performance of BERT and EL\_MultiBERT models for each named entity class. As can be seen from the table, our entity linking approach improves the performance, especially in complex entities like \textit{Creative Work}, \textit{Group} and \textit{Product}. Two approaches returned similar results only for \textit{Person} and \textit{Location}, but since we do not have the gold standard labels, we cannot perform any detailed analysis of the reason.  

\subsection{Ensemble Approach}

Although the majority of the samples in the test set are short texts, there are some longer sentences as well. From the validation set experiments, we already know that when there is enough context, transformer models like BERT perform quite well. Hence, we decided to see if we can use our proposed architecture as a fall back mechanism when there is not enough context available, but otherwise use BERT.  
We applied an ensemble approach relying on BERT for samples with more than 11 tokens and EL\_MultiBERT for shorter ones. So when sentence length is less than 11, we depended on our entity linking architecture and Wikipedia for retrieving and using additional useful context about the input. 
As shown in Table 3, The ensemble of BERT and EL\_MultiBERT achieved the best result with 72.02\% on the test set, and we ranked as third in the Turkish track.


\section{Conclusion}

In this work, we proposed a language-independent method for the Complex NER task in low-context settings. Our results showed that utilizing the use of entity linking for NER provides a significant improvement over state-of-the-art transformer models when there is not enough context. Yet, since our resources are limited, we experimented with only the Turkish part. In future work, we will extend this approach in a multilingual setting.



\bibliography{semeval22}
\bibliographystyle{acl_natbib}

\end{document}